\pdfoutput=1

\documentclass[11pt]{article}

\usepackage{EMNLP2023}

\usepackage{times}
\usepackage{latexsym}

\usepackage[T1]{fontenc}

\usepackage[utf8]{inputenc}

\usepackage{microtype}

\usepackage{inconsolata}

\usepackage{times}
\usepackage{amsmath}
\usepackage{algorithm}
\usepackage{algpseudocode}
\usepackage{diagbox}
\usepackage{latexsym}
\usepackage{enumitem}
\usepackage{siunitx}

\usepackage{helvet}  
\usepackage{courier}  
\usepackage{url}  
\usepackage{graphicx}  
\usepackage{courier}
\usepackage{amsmath}
\usepackage[normalem]{ulem}
\useunder{\uline}{\ul}{}
\usepackage{graphicx}
\usepackage{multirow}
\usepackage{mathbbol}
\usepackage{caption}
\usepackage{CJKutf8}
\usepackage{verbatim}
\usepackage{float}
\usepackage{booktabs}
\usepackage{array}
\usepackage{times}
\usepackage{csquotes}
\usepackage{latexsym}
\usepackage{tablefootnote}
\usepackage[normalem]{ulem}
\usepackage{todonotes}
\usepackage{subcaption}
\usepackage{tikz}
\usepackage{xcolor}
\usepackage{hhline}

\title{Improved Beam Search for Hallucination Mitigation in Abstractive Summarization}

\author{Arvind Krishna Sridhar, Erik Visser \\
  Qualcomm Technologies \\
  {\tt \{arvisrid, evisser\}@qti.qualcomm.com} \\}

\begin{document}
\maketitle
\begin{abstract}
 Advancement in large pretrained language models has significantly improved their performance for conditional language generation tasks including summarization albeit with hallucinations. With the rise in the commercial use of text-generative applications, it has become necessary to have a component that ensures the factuality of the responses. To reduce hallucinations, conventional methods proposed improving beam search or using a fact checker as a postprocessing step. In this paper, we investigate using the Natural Language Inference (NLI) entailment metric to detect and prevent hallucinations in summary generation. We propose an inference time and easily generalizable NLI-assisted beam re-ranking mechanism by computing entailment probability scores between the input context and summarization model-generated beams during saliency-enhanced greedy decoding. We also investigate the limitations of existing academic factuality benchmarks and demonstrate that our proposed algorithm consistently outperforms the baselines in human evaluation on publicly available XSum and CNN/DM datasets.
\end{abstract}

\section{Introduction}\label{sec:intro}
Pretrained sequence-to-sequence transformer-based models like BART \cite{lewis2019bart}, Pegasus \cite{Zhang2020Pegasus} etc. have shown substantial improvements in the performance of NLP tasks like summarization, story generation, abstractive question answering, etc. Hallucination is a common issue observed during the generation process  \cite{Ji2022SurveyHallucination} as the pretraining is largely conducted on unlabeled data \cite{lewis2019bart}. During this pretraining phase, the model learns the inaccuracies of training data along with its grammar and often generates words that are not pertinent to the given input during inference time.  

Research has been conducted at curbing hallucination during the decoding phase. \cite{King2022} proposed a modification to beam search by constraining the decoding step to focus on input-supported tokens. They hypothesize that the inaccuracies in gold summaries give rise to inconsistencies in the generated text. \cite{xu-etal-2020-understanding-neural} and \cite{van-der-poel-etal-2022-mutual} investigated the relationship between hallucination and predictive uncertainty and proposed an extension to beam search by preferring low predictive uncertainty.

While there has been some success in constraining beam search using heuristics functions, they require manual inspection using intricate knowledge of the dataset, task and model to initialize their hyperparameters. This isn't feasible in a production system where the test time data distribution can have significant variations over time. PINOCCHIO \cite{King2022} uses cosine distance to measure the consistency of generated word with context at each decoding step. As the dataset becomes more abstractive, relying solely on cosine distance and simple word-level heuristics is ineffective in steering the beam decoding factually. \cite{balachandran-etal-2022-correcting} proposed a Bart-based fact correction model that needs to be fine-tuned specifically for each dataset and adds to the overall model complexity.

Our proposed approach overcomes the limitations of heuristics and extra parameter complexity. To our knowledge, we are the first to introduce the semantically matching NLP task of Natural Language Inference (NLI) during decoding phase to re-rank the top few predictions of the model. Previous studies used NLI-trained models to re-rank summaries like \cite{falke-etal-2019-ranking, Mitchell2022EnhancingUsingNLI} and compared the complete summary candidates with context. They empirically show that NLI entailment probability alone isn’t enough to differentiate the correct summary beams from incorrect ones.

In this work, we compute NLI entailment scores at beam decoding step to provide the model an opportunity to change the beam track towards a less hallucinated region. Each intermediate beam is generated using greedy rollout decoding while attending to salient context parts. Then, the beams are ranked using the NLI metric. We make the following contributions:

1) We develop a hallucination mitigation component for beam search that can modify the cumulative beam probability at the token level using the NLI metric.\\
2) We showcase the limitations of the existing state-of-the-art factuality metrics including FactCC, SummaC and QGQA scores. We provide demonstration examples in \ref{sec:appendix} to strengthen our case.\\
3) Using human evaluation, we demonstrate the effectiveness of our proposed approach against the baselines.

\section{Related Work}
\subsection{Measuring and improving faithfulness}
Faithfulness refers to how consistent the generated text is with respect to the input. Throughout this paper, we use the term factually inconsistent to be synonymous with hallucinated text.  \cite{maynez-etal-2020-faithfulness} assessed the types of hallucinations produced by different abstractive summarizers (RNN-based Seq2Seq \cite{Abigail2017Pointer}, GPT-tuned \cite{Radford2019LanguageMA}, BertS2S \cite{devlin2018bert}).

To measure factual inconsistency,  \cite{kryscinski-etal-2020-evaluating} trained FactCC – a Bart-Base model finetuned on synthetically hallucinated summaries using semantically variant and invariant transformations like Entity Swap, Sentence Negation, Paraphrasing and Noise Injection. But such a black box model lacks interpretation, has low generalizability to other datasets and is only adept at finding minor hallucinations like the transformations. \cite{durmus-etal-2020-feqa, fabbri-etal-2022-qafacteval} have explored QA based metrics to measure factual inconsistencies by generating and comparing question-answer pairs across generated and ground truth summaries.

To generate faithful summaries, numerous architecture modifications have been proposed \cite{10.1145/3511808.3557319, zhang-etal-2022-improving-faithfulness}. Such modifications are neither generalizable nor feasible to be incorporated into industry text generation models with various constraints and train their existing summarizer models from scratch. Research has also been into improving the loss function component to improve overall factual accuracy. For example,  \cite{kang-hashimoto-2020-improved} demonstrated that the model shows increased factual accuracies by truncating the loss by adaptively removing high log loss examples. 

\subsection{Hallucinations in abstractive summarization}
 \cite{Ji2022SurveyHallucination} documented an extensive survey on hallucinations present in various NLP downstream tasks. They discuss the different hallucination mitigation methods with respect to these tasks and summarize the metrics to measure hallucination. An abstract summary is defined to be hallucinated if it has any spans of text not semantically supported by the input document. Hallucinations can be categorized into two major types – intrinsic and extrinsic. Intrinsic hallucinations refer to the contradictions in the abstract summary with respect to the input document: for example, using wrong pronouns, swapping names and verbs. Models like FactCC \cite{kryscinski-etal-2020-evaluating} trained on minor text transformations can be used to detect such errors. Extrinsic hallucinations refer to the unsupported spans of text present in the generated summaries that cannot be verified only with the input document. It arises partly due to the extrinsic hallucinations present in human written summaries which the model overfits during training process.  \cite{dziri-etal-2022-origin} studied hallucinations in conversational models using their corresponding datasets and observed that Sequence-2-Sequence models like GPT-2 not only hallucinate but also amplify the percentage of hallucinations with respect to the training data. \cite{cao-etal-2022-hallucinated, dong-etal-2022-faithful} inspect whether the hallucinations generated in summary align with world knowledge. 

\section{Methodology}\label{sec:model}

\begin{figure*}[t!]
\centering
\includegraphics[width=1.0\textwidth]{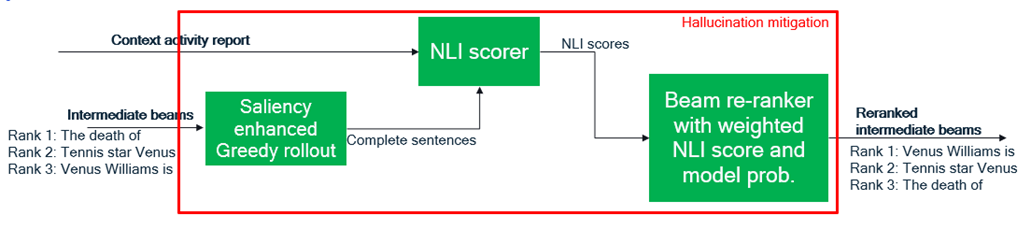}
\caption{Proposed beam search re-ranker}
\label{fig:beamreranker}
\vspace*{-3mm}
\end{figure*}

In this section, we describe the components of our NLI-aided beam search re-ranker as shown in Figure \ref{fig:beamreranker}. For all experiments, we use a BART-Base model \cite{lewis2019bart} finetuned with the given dataset. Architectures like BART \cite{lewis2019bart} have an autoregressive decoder that generates the output word by word conditioned on the input text and the words generated so far as shown in Equation \ref{eq:1}. As mentioned in  \cite{meister-etal-2020-beam}, Beam search performs a breadth-first search with limited branches with the beam size starting with the BOS token(Begin of sentence) and ending the search at EOS token(End of sentence). Each path from BOS to EOS is called a hypothesis.  

We define intermediate beam or partial hypothesis as the sequence of sub paths of hypotheses starting at BOS and ending before EOS. The saliency enhanced greedy rollout component, explained in Section \ref{sec:4.1}, attends over important parts of the context relevant to the intermediate beams and completes the beam till EOS.

The intermediate beams are sent to the saliency-enhanced greedy rollout to serve as a look-ahead mechanism to complete the beams. The completed candidate beams are scored using the entailment probabilities from the NLI model. Then the intermediate beams are re-ranked based on the weighted probabilities between entailment and the model probabilities. Detailed steps of our proposed algorithm are provided in Algorithm \ref{algorithm:1}. We adopt required variable names from \cite{King2022} for consistency.


\begin{equation}
\vspace*{-2mm}
P_\Theta (y|x) = \prod_{t=1}^{|y|}P_\Theta (y_t | x, y_{<t}),
\label{eq:1}
\end{equation}

\subsection{Saliency-enhanced Greedy rollout}\label{sec:4.1}

Since it is difficult to perform the NLI task on partial hypotheses as the NLI models have been trained with complete sentences \cite{maccartney-manning-2008-modeling}, we complete 2B intermediate beams as our first step where B is the beam size. Inspired by  \cite{hargreaves-etal-2021-incremental}, we use the greedy search on the intermediate beams to generate the remaining words and complete the partial hypotheses. In Algorithm \ref{algorithm:1}, the saliency-enhanced greedy rollout (SGR) function takes the concatenated input of context, the intermediate beam and the next word separated by [SEP] token and generates the completed beams. During the greedy search, we empirically witnessed similar words being used to complete the beams regardless of the words in intermediate beams. This might be due to the long context and shorter attention span of pretrained transformers as mentioned in  \cite{liu-lapata-2019-text}. Thus the model might not effectively attend to the parts of context relevant to the words in the intermediate beam. To solve this problem, we take two steps. 

First, inspired by  \cite{Cao2021AttentionHM}, we enhance the effectiveness and diversity of greedy search, by introducing saliency on the context relative to the intermediate beam using attention head masking. We compute the saliency score for every word in context by averaging its cosine distance with each word in the intermediate beam. We propose two saliency versions v1 and v2 suitable for summaries with extractive and abstractive characteristics respectively. In Saliency v1, using a threshold as a hyperparameter, we compute mask matrix m (Equation \ref{eq:2}) to selectively attend to words in the context relevant to the completion of the current intermediate beam. A hard masking is done so that it increases the probability of copying relevant words from the input. But in a highly abstractive summarization setting, attending over all words is crucial. Hence, we propose Saliency v2 that performs variable soft attention over the words in the input. In Saliency v2, inspired by GATE \cite{ahmad2021gate}, we use the computed saliency scores to perform weighted attention (Equation \ref{eq:3}) on the softmax output and normalize the scores(Equation \ref{eq:4}). 

Second, we perform the proposed re-ranking only if the hypothesis has a predefined minimum of words so that the beam doesn’t converge to the same space during greedy search. This is because if the hypothesis has few words, the beam might not have the necessary entities suitable for measuring hallucination. In the future, we can identify the appropriate time steps suitable for re-ranking the hypothesis to avoid hallucinations.


\begin{equation}
\vspace*{-2mm}
\textrm{Attention(q,K,V)} = \textrm{softmax}(\frac{qK^T}{\sqrt{d_k}} + \textrm{m}) \textrm{V}
\label{eq:2}
\end{equation}
Where q is query, K and V represent key and value matrices respectively, $d_k$ is the scaling factor and 
m is the attention mask matrix.

\begin{equation}
\vspace*{-2mm}
\textrm{Attention(q,K,V)} = \textrm{F}(\textrm{softmax}(\frac{qK^T}{\sqrt{d_k}} + \textrm{m}) \textrm{V})
\label{eq:3}
\end{equation}

\begin{equation}
\vspace*{-2mm}
{\textrm{F(P)}_{ij}} = \frac{\textrm{P}_{ij}}{\textrm{Z}_{i}\textrm{D}_{ij}}
\label{eq:4}
\end{equation}
Where $Z_i = \sum_{j}\frac{P_{ij}}{D_{ij}}$ is the normalization factor and $D_{ij}$ is the saliency score between ith token in intermediate beam and jth token in document.

\begin{algorithm*}
   \caption{NLI Assisted Beam Decoding}
   \small
    \begin{algorithmic}
      \State \textbf{Input variables: }{Beam size B, Generative Model M, Vocab V, wait threshold $\mathbf{\delta}$} \\
      \textbf{Input functions: }{Saliency enhanced greedy rollout function $\mathbf{SGR}$ returns completed beams, \\
 Natural Language Inference function $\mathbf{NLI}$ returns entailment probability; }\\
      \textbf{Initialize: }{I = \{($\textrm{Inter}_i$, Cumulative $\textrm{P}_i$) : $\textrm{Inter}_i$ $\in$ set of Intermediate beams\} };\\
       Context C = \{$x_1, x_2, x_3,….x_n : x_i$ $\in$ V \}; Candidate Intermediate beams CI = \{\};\\
        Current Completed beams CC = \{\}; Completed Generations CG = \{\};\\
      \textbf{Output: }{ top-ranked elements of CG}

        \While{|CG| < B}
            \For{($\textrm{Inter}_i$, Cumulative $\textrm{P}_i$) in I}
                 \State T := \{ ($\textrm{t}_i$, $\textrm{P}_i): \textrm{t}_i$ $\in$ Top 2B tokens of V predicted by Model M with $\textrm{P}_i$ probability\}
                \If{|$\textrm{Inter}_i$| > $\delta$ }
                    \For{($\textrm{t}_i$, $\textrm{P}_i$) in T}
                         \State R:= $\mathbf{SGR}$(M, C, $\textrm{Inter}_i$, $\textrm{t}_i$)
                         \State $\textrm{P}_{entail}$  := $\mathbf{NLI}$(C, R)
				         \State $\textrm{P}_{weighted}$ := $\alpha\textrm{P}_i$  + (1-$\alpha$) $\textrm{P}_{entail}$
                         \State $\textrm{P}_{i}$ := $\textrm{P}_{weighted}$
                     \EndFor
                 \EndIf
             \State CI := \{($\textrm{Inter}_i$ + $\textrm{t}_{i}$, Cumulative $\textrm{P}_i$ + $\textrm{P}_i$): ($\textrm{t}_{i}$, $\textrm{P}_{i}$) $\in$ T\}
		   \State I := I U Top B beams from CI ranked by Cumulative P 
		   \State CC:= \{$\textrm{Inter}_i$ : for all beams $\textrm{Inter}_i \in$ I ending with '<end>' token\} 
		   \State I:= I - CC
		    \State CG:= CG U CC
            \EndFor
        \EndWhile \\
         \Return{top ranked elements of CG}

\end{algorithmic}
\label{algorithm:1}
\vspace*{-1mm}
\end{algorithm*}
\vspace*{-1mm}

\subsection{Natural Language Inference(NLI) scorer}\label{sec:NLI scorer}
As a next step, we pass the greedy rollout beams to the NLI scorer. We obtain the entailment probability with the context as premise and the beam as hypothesis \cite{maccartney-manning-2008-modeling} as illustrated with Equation \ref{eq:4}. The NLI function takes in Context C as the premise and rolled out beam R as the hypothesis and computes their relationship as entailment score. We hypothesize that the entailment probability is inversely proportional to the hallucination content of the beam. 

\subsection{Weighted Beam re-ranker}\label{sec:4.3}
In order to incorporate the NLI score into the overall cumulative beam probability, we take a weighted average of entailment and model probabilities for each decoding step and add them to the cumulative beam probability. We then re-rank the beams based on the modified cumulative probability and select the top $B$ candidates as re-ranked intermediate beams. The weights need to be normalized as we are adding two random variables. As mentioned in Equation \ref{eq:4}, we consider weight($\alpha$) as a hyperparameter which can be increased up to 1.0 depending on the necessity of faithfulness in the generated text for a given task.

\begin{align*}
\vspace*{-2mm}
\textrm{P}_{entail}  &:= \textrm{NLI(C, R)} \\
\textrm{P}_{weighted} &:= (1-\alpha) \textrm{P}_i  + \alpha \textrm{P}_{entail} \label{eq:4} \tag{4}
\end{align*} \\
Where NLI is the Natural Language Inference function taking Context C, and Intermediate beam Rollout R as inputs.
\section{Experiments}

\subsection{Dataset}\label{sec:dataset}
We use two datasets, namely, CNN/DM \cite{Karl2015TeachingMachines} and XSum \cite{narayan-etal-2018}, to evaluate our model performance. CNN/DM corpus is generated from human-written multi-line summaries for the CNN and Daily Mail news articles. It consists of over 285k training pairs, 13,368 validation pairs and 11,487 test pairs. The XSum dataset is made up of BBC articles and their one-line summaries. It comprises over 90k training samples and is more abstractive than CNN/DM as it contains 18.6\% more novel unigrams. We aim to develop an approach that works consistently on both abstractive and extractive types of summaries.

\subsection{Evaluation Metric}\label{sec:evaluationmetric}
We use multiple factuality metrics including SummaC-Conv, SummaC-CZS \cite{laban-etal-2022-summac}, QGQA \cite{fischer2022measuring} and FactCC \cite{kryscinski-etal-2020-evaluating} to showcase the relative performance of baselines and our approach. FactCC is a binary classifier for factuality. Since we are comparing models against referenceless metrics, we consider the probability of not hallucinated as FactCC score. For QGQA, we use its F1 score. We penalize the cases when the algorithm doesn't generate a summary with corresponding low score. We provide a comprehensive human evaluation study consisting of 50 instances randomly drawn from the XSum and CNNDM datasets equally. 4 human annotators with more than 28 years of combined professional and academic English proficiency scored between 1(lowest) to 5(highest) to evaluate the faithfulness of summaries. 
\section{Results}\label{sec:result}
We demonstrate that our proposed beam search modification reduces hallucination during inference time by comparing it with multiple baselines including vanilla beam search, PINNOCHIO \cite{King2022} and FactEdit \cite{balachandran-etal-2022-correcting}. From Table \ref{Table:mainresult}, we can observe contrasting trends for both the datasets. For Xsum, our approach performs better than the baselines except for QGQA. While for CNN/DM, the metrics tend to favour FactEdit. We hypothesize that this behavior is due to Vanilla beam search and FactEdit algorithms tending to generate slightly more extractive summaries in comparison to the other methods. Similar to the conclusion in a contemporary paper \cite{tang-etal-2023-understanding}, we also observe that none of the factuality metrics are consistent across datasets for the Bart-based summarization model. In Table \ref{Table:humanevaluation}, We perform human evaluation to rate the summaries subjectively. The high faithfulness score of the proposed approach compared to other baselines bolsters the efficacy of our algorithm.

\begin{table*}[t!]
\begin{center}
\small
\centering
\begin{tabular}{|p{1.2cm}|p{2.8cm}|p{2.1cm}|p{1.8cm}| p{0.9cm}| p{0.9cm}| }
\hline
\textbf{Dataset} & \textbf{Decoding Algorithm}   & \textbf{SummaC-Conv} & \textbf{ SummaC-ZS} & \textbf{FactCC} & \textbf{QGQA}\\ \hline
XSum & Vanilla beam search & 0.244 & -0.358 & 0.248 & \textbf{0.843} \\ \hline
& FactEdit & 0.245 & -0.356 & 0.249 & 0.837 \\ \hline
& PINNOCHIO & 0.219 & -0.4022 & 0.205 &  0.786\\ \hline
& Ours (Saliency v1) & \textbf{0.248} & -0.289  & \textbf{0.280} & 0.841 \\ \hline
& Ours (Saliency v2 & 0.248 & \textbf{-0.279} & 0.219 & 0.839 \\ \hhline{|=|=|=|=|=|=|}

CNN/DM & Vanilla beam search & 0.524 & 0.137 & 0.352 & \textbf{0.948} \\ \hline
& FactEdit & 0.536 & \textbf{0.142} & \textbf{0.412} & 0.945 \\ \hline
& PINNOCHIO & \textbf{0.676} & 0.141 & 0.344 & 0.942\\ \hline
& Ours (Saliency v1) & 0.449 & 0.050 & 0.212 & 0.930 \\ \hline
& Ours (Saliency v2 & 0.439 & 0.032 & 0.219 & 0.932 \\ \hline

\end{tabular}
\end{center}
\caption{Performance of baselines and our proposed approach on XSum and CNN/DM datasets.}
\label{Table:mainresult}
\end{table*}

\section{Analysis}

In Table \ref{table:alphaexample}, we investigate the role of hyperparameter $\alpha$ in guiding the beams to factual generation by varying its values across the spectrum. For $\alpha$ = 0.0, all the generated beams are factually wrong. While for $\alpha$ = 0.2 and 0.8 the generated beams are factually consistent. The beams with non-zero $\alpha$ are inherently more diverse than the vanilla beam search generations.

To quantify whether the beams were able to explore different regions of text space, we propose a diversity metric (Equation \ref{eq:5}) to measure the average frequency of novel words across the beams. In the future, we plan to evolve the set intersection operation to incorporate a semantic representation of words. 
\begin{equation}
\vspace*{-2mm}
\textrm{Diversity} = \frac{\sum_{i=0}^{n-2}b_i \bigcup b_{i+1} - b_i \bigcap b_{i+1}}{_{2}^{n}\textrm{C}}
\label{eq:5}
\end{equation}
Where n is the beam size, $b_i$ is the set of unique words in beam $i$.
\begin{table*}[t!]
\small
\centering
\begin{tabular}{|p{2.2cm}||p{14cm}|}
\hline

\textbf{Gold Summary} & US tennis star Venus Williams has been involved in a car accident that led to the death of a 78-year-old man.\\ \hline
\end{tabular}

\begin{tabular}{|p{2.2cm}||p{14cm} |}
\hline
\textbf{$\alpha$} & \textbf{Generated Summary} \\ \hline
\end{tabular}
\begin{tabular}{|p{2.2cm}||p{14cm}|}
\hline
  & \colorbox{red}{Tennis star Venus Williams has died in a car crash in Florida, police say.} \\ 
  &  \colorbox{red}{Tennis star Venus Williams has died in a car crash in Florida, US police say.} \\ 
 0.0 & \colorbox{red}{Tennis star Venus Williams has died in a car crash in Florida, according to reports.} \\     
  & \colorbox{red}{Tennis star Venus Williams has died in a car crash in Florida, according to police.} \\ 
  &  \colorbox{red}{Tennis star Venus Williams has died in a car crash in Florida.}\\ 
\hline
\end{tabular}

\begin{tabular}{|p{2.2cm}||p{14cm}|}
\hline
  &  Tennis star Venus Williams was at fault for a car crash that killed a man in Florida, police say. \\ 
  & Tennis star Venus Williams was at fault for a crash that killed a man in Florida, police say. \\ 
 0.2 & Tennis star Venus Williams is being investigated over the death of a man in Florida, police say.\\     
  & Tennis star Venus Williams was at fault for a car crash that killed a man, police say. \\ 
  &  Tennis star Venus Williams is being investigated over the death of a man in a car crash in Florida.\\ 
\hline
\end{tabular}

\begin{tabular}{|p{2.2cm}||p{14cm}|}
\hline
  &  Tennis star Venus Williams is being investigated over the death of a man in Florida, police say. \\ 
  & Tennis star Venus Williams was at fault for a car crash that killed a man in Florida, according to reports. \\ 
 0.8 & Tennis star Venus Williams was at fault for a car crash that killed a man in Florida, police say.  \\     
  & Tennis star Venus Williams was at fault for a car crash that killed a man in Florida, according to police. \\ 
  & Tennis star Venus Williams is being investigated for causing the death of a man by careless driving, police say. \\ 
\hline
\end{tabular}
\caption{An example, from XSum dataset, illustrating the effect of hyperparameter $\alpha$ for beam size 5. Highlights in Red indicate factually inconsistent beams.}
\label{table:alphaexample}
\end{table*}

In Table \ref{Table:diversityresult}, we evaluate the diversity score for vanilla beam search and the proposed approach variations. Since, XSum has more chances of hallucination due to its abstractive nature, a higher diversity score would enable the algorithm to explore more regions for reducing hallucination. Whereas, CNN/DM being highly extractive, doesn't require much re-ranking during the decoding and hence the low diversity.

\begin{table}[t!]
\begin{center}
\small
\centering
\begin{tabular}{|p{2.8cm}|p{1cm}|p{1.2cm}| }
\hline
 \textbf{Decoding Algorithm}  & \textbf{XSum} & \textbf{CNN/DM}\\ \hline
Vanilla beam search & 2.27 & 1.75 \\ \hline
Ours (Saliency v1) & 2.53 & 1.04 \\ \hline
Ours (Saliency v2) & 2.51 & 1.01 \\ \hline
\end{tabular}
\end{center}
\caption{Diversity scores of vanilla beam search and proposed approach.}
\label{Table:diversityresult}
\end{table}

\begin{table}[t!]
\begin{center}
\small
\centering
\begin{tabular}{|p{2.8cm} | p{1.6cm}| }
\hline
 \textbf{Decoding Algorithm}  & \textbf{Faithfulness} \\ \hline
Vanilla beam search & 3.37  \\ \hline
Pinnochio & 3.21 \\ \hline
FactEdit & 3.32 \\ \hline
Ours (Saliency v2) & \textbf{3.48}*  \\ \hline
\end{tabular}
\end{center}
\caption{Human Evaluation on faithfulness metric for baselines and proposed approaches. * denotes statistically significant over Pinnochio and FactEdit with 95\% confidence interval on paired-t test. }
\label{Table:humanevaluation}
\end{table}

\begin{table}[t!]
\begin{center}
\small
\centering
\begin{tabular}{|p{2.5cm}|c|c|c|c|}
\hline
 \multirow{2}{2.5cm}{\textbf{Rollout decoding vs Dataset}}  & \multicolumn{2}{c|}{\textbf{XSum}} & \multicolumn{2}{c|}{\textbf{CNN/DM}} \\
     \cline{2-5}
    & \textbf{S-Conv} & \textbf{S-ZS}  & \textbf{S-Conv}  & \textbf{S-ZS} \\
 \hline
 
Random Sampling & 0.247 & -0.299 & 0.599 & 0.042 \\ \hline
Top K  & 0.246 & \textbf{-0.279} & \textbf{0.605} & \textbf{0.049}  \\ \hline
Top P & 0.247 & -0.291 & 0.601 & 0.047 \\ \hline
Greedy & \textbf{0.248} & -0.289 &  \textbf{0.605} & \textbf{0.049}  \\ \hline
\end{tabular}
\end{center}
\caption{Analysis of different decoding strategies for rollout component. S-Conv and S-ZS stands for SummaC-Conv and SummaC-ZS.}
\label{Table:rollout}
\end{table}

Next, we inspect the role of the decoding algorithm during the rollout of intermediate beams on the overall performance of the algorithm. From Table \ref{Table:rollout}, We can see that Top K and Greedy search perform superior to their counterparts and can be concluded as better lookaheads for faithful decoding. Please refer to Appendix \ref{sec:appendix}, for example demonstrations and further analysis.

\section{Conclusion and Future Work}
We propose a modification to the beam search decoding algorithm that guides beam generation to avoid falling into hallucination regions by re-ranking the beams based on NLI entailment scores computed on saliency-enhanced greedily rolled-out partial hypotheses. We propose two variations of saliency v1 and v2 appropriate for extractive and abstraction style summarisation settings. We present the issue of inconsistency of SOTA Summarization factuality metrics to motivate the development of a robust benchmark for detecting hallucinations. By human evaluation, we show that our NLI-based re-ranker consistently improves the faithfulness score. In the future, we intend to investigate NLI-based re-ranker’s performance on other NLP downstream tasks such as story generation with prompt, question answering and query-focused summarization. Also, we plan to study more efficient methods to incorporate the NLI as a guidance mechanism for decoding algorithms. We also intend to study the efficacy of our approach on tasks with longer context like long document summarization usinf a retrieval augmented ranker.

\section*{Limitations}
The re-ranking of intermediate beams at regular intervals might introduce some delay in the decoding process. We show in Table \ref{Table:rerankingxsum} and \ref{Table:rerankingcnndm} (Appendix \ref{sec:appendix}), that for smaller variations in re-ranking interval, the beams still follow the same faithfulness guided path. Thus, we believe that such latency could be offset by setting the appropriate interval size and parallel processing of the beams during inference time decoding.

\section*{Ethics Statement}
We believe our work has no negative ethical impact on society and strongly hope our NLI-based reranker help in creating a positive impact by promoting faithfulness among text generative models conditioned on input text specifically commercial production running systems. Our work will aid in reducing false and biased information generated by LLMs.
Our NLI and BART models are trained on open source datasets, CNNDM and XSum, which are said to contain hallucinations and biases from the news media. Whenever possible, it is recommended to retrain an unbiased NLI model.

\bibliography{anthology,custom}
\bibliographystyle{acl_natbib}

\appendix
\section{Appendix}
\label{sec:appendix}
\subsection{Implementation Details}\label{sec:implementationdetail}
We used the PyTorch \cite{NEURIPS2019_bdbca288} implementation of the BART base version from hugging face library \cite{wolf-etal-2020-transformers}. We train for 6 epochs using a learning rate of 4e-5 with linear decay. For the decoding process, we use beam search with beam size 5 and a maximum length of 125 tokens after Byte Pair Encoding(BPE) tokenization. Early stopping is used while the repetition penalty is 1.0 and 3.0 for XSum and CNN/DM datasets respectively. For NLI, we use BART-Large model finetuned on MNLI dataset \cite{williams-etal-2018-broad}. Unless explicitly specified, we use $\alpha$ = 0.8 for both the datasets. Due to resource constraints, we perform the evaluation on 820 randomly selected test samples from each of the datasets. For the rollout generation, we constrain it to a max length of 60 tokens.

\subsection{Using NLI to detect hallucinations}
 \cite{maccartney-manning-2008-modeling} defines Natural Language Inference as the task of determining whether a natural-language hypothesis can be inferred from a given premise. Given a premise and a hypothesis, NLI computes the relationship between them in the form of three probabilities – entailment, contradiction and neutral. For Algorithm \ref{algorithm:1}, we mainly consider the entailment score. In this section, we use the recognizing textual entailment task as defined in  \cite{maccartney-manning-2008-modeling} to detect hallucinations in abstractive summarization task. Intrinsic hallucinations are harder to detect as they require more than lexical matching to deduce the relevance of a given word with context. To quantify the hallucinations, we consider only entity-based hallucinations for the purpose of analysis.

\begin{figure}[t!]
\centering
\includegraphics[width=0.5\textwidth]{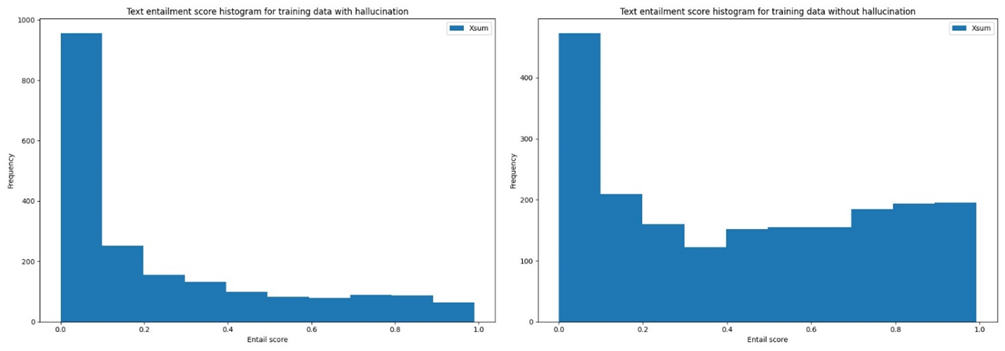}
\caption{Histogram of entailment scores for the XSum training data (a) with and (b) without hallucinations}
\label{fig:nlientailment}
\vspace*{-1mm}
\end{figure}

\begin{table}[!htbp]
\centering
\begin{tabular}{|l|l|l| }
\hline
\textbf{Dataset} & \textbf{Hallucinated}  & \textbf{Not Hallucinated}\\ \hline
XSum & 0.243 & \textbf{0.433} \\ \hline
\end{tabular}
\caption{Average entailment scores of the XSum training data on 2000 samples}
\label{Table:nlientail}
\vspace*{0.1mm}
\end{table}

NLI models are fine-tuned to detect the relationship between single sentence premise and hypothesis. In order to verify whether their classification can extend to multisentence premise and single sentence hypothesis setting, we conducted an experiment to analyze the correlation between entailment scores and entity hallucinations on randomly selected 2000 training samples in the XSum dataset. It is crucial to note that Xsum is a single sentence summary dataset and . Hence, we effectively compare a single sentence against an array of sentences in the document. From Figure \ref{fig:nlientailment}, it is evident that although there is a high frequency of low entailment scores for both data with and without hallucinations, the distinction between them becomes clearer at higher entailment scores. Indeed a higher entailment score correlates with low probability for entity hallucinations. This is also reflected in the average entailment scores as shown in Table \ref{Table:nlientail}. This analysis proves that using a multi-sentence premise, the NLI measure can detect entity-based hallucination in a single-sentence hypothesis. We use this argument and setting to introduce NLI during the beam decoding.

\subsection{Analysis}
\begin{figure}[hbt!]
  \centering
  \subfloat[XSum]{\includegraphics[width=0.3\textwidth]{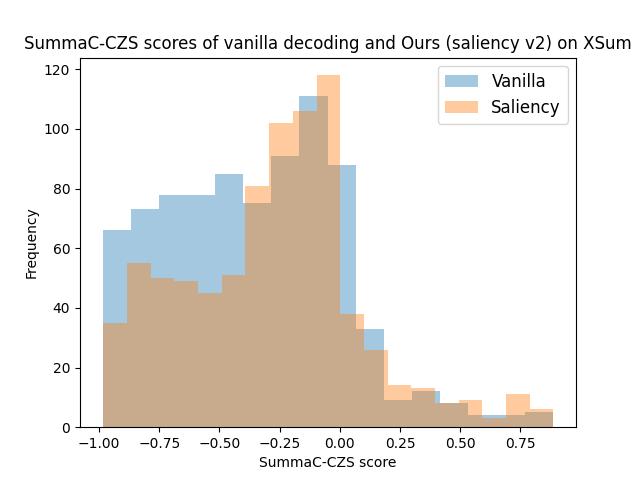}\label{fig:f1}}
  \hfill
  \subfloat[CNNDM]{\includegraphics[width=0.3\textwidth]{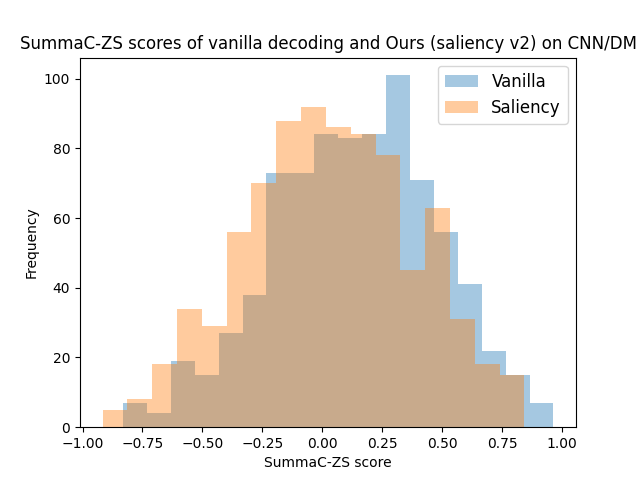}\label{fig:f2}}
  \caption{SummaC-ZS distribution of vanilla beam search and Ours (Saliency v2)}
  \label{fig:distributionCZS}
\end{figure}

We perform a few additional ablation studies and analysis to understand our algorithm in depth. In Figure \ref{fig:distributionCZS}, we visualize the SummaC-ZS scores of vanilla beam search and Ours (saliency v2). In CNNDM dataset, the similar score distribution of vanilla beam search and our algorithm shows that a few outlier scores, especially in 0.25 to 0.50 bucket, have a high influence on the mean SummaC-CZS scores. We suspect this might be a possible explanation for the low performance of our algorithm on CNNDM dataset.

Next, we investigate whether our algorithm is able to guide the beam search towards faithful regions even if we increase the re-ranking interval i.e. performing the NLI-based re-ranking once for x number of tokens. From Table \ref{Table:rerankingxsum} and \ref{Table:rerankingcnndm}, it is evident that the beams do not fall off the right track even if we perform re-ranking at slightly longer intervals. This result is particularly important as we can tweak and increase re-ranking interval to achieve lower inference latency.
\begin{table}[!htbp]
\begin{center}
\small
\centering
\renewcommand{\arraystretch}{1.2}
\begin{tabular}{|p{2.5cm}|c|c|c|c|}
\hline
Re-ranking Interval & S-Conv & S-CZS & QGQA \\ \hline
2 & 0.2453 & -0.3402 & 0.8403 \\ \hline
4 & 0.2453 & -0.3402 & 0.8403 \\ \hline
8 & 0.2453 & -0.3402 & 0.8403 \\ \hline
\end{tabular}
\end{center}
\caption{Effect of re-ranking interval on overall performance for Xsum dataset.  S-Conv and S-ZS stands for SummaC-Conv and SummaC-ZS.}
\label{Table:rerankingxsum}
\end{table}

\begin{table}[!htbp]
\begin{center}
\small
\centering
\renewcommand{\arraystretch}{1.2}
\begin{tabular}{|p{2.5cm}|c|c|c|c|}
\hline
Re-ranking Interval & S-Conv & S-CZS & QGQA \\ \hline
16 & 0.605 & 0.049 & 0.930 \\ \hline
32 & 0.605 & 0.049 & 0.930 \\ \hline
\end{tabular}
\end{center}
\caption{Effect of re-ranking interval on overall performance for CNNDM dataset. S-Conv and S-ZS stands for SummaC-Conv and SummaC-ZS.}
\label{Table:rerankingcnndm}
\end{table}

\begin{table*}[t!]
\begin{center}
\small
\centering
\renewcommand{\arraystretch}{1.2}
\begin{tabular}{|p{2.5cm}|c|c|c|c|c|c|c|c|c|}
\hline
    \multirow{2}{3cm}{\textbf{Decoding Algorithm vs Dataset}} & \multicolumn{4}{c|}{\textbf{XSum}} & \multicolumn{4}{c|}{\textbf{CNN/DM}} \\
    \cline{2-9}
    & \textbf{Diversity} & \textbf{S-Conv} & \textbf{S-ZS}  & \textbf{QGQA} & \textbf{Diversity} & \textbf{S-Conv} & \textbf{S-ZS}  & \textbf{QGQA}  \\
\hline

Vanilla beam search & 2.27 & 0.244 & -0.358 & \textbf{0.843} & 1.75 & \textbf{0.683} & \textbf{0.137} & \textbf{0.948}\\ \hline
Ours : E & 2.54 & \textbf{0.248} & -0.289 & 0.841 & 1.04 & 0.605 & 0.049 & 0.930 \\ \hline
Ours : E \& C & 2.54 & 0.247 & \textbf{-0.270} & 0.841 & 0.99 & 0.604 & 0.053 & 0.930 \\ \hline
\end{tabular}
\end{center}
\caption{Effect of \textbf{E}ntailment and \textbf{C}ontradiction NLI probabilities on overall performance}
\label{Table:contratable}
\end{table*}

Finally, we study if utilizing both entailment and contradiction probabilities aids in re-ranking the beams. $\textrm{P}_{weighted}$ in Algorithm \ref{algorithm:1} will be modified using Equation \ref{eq:weightedcontra}. For this experiment, we assigned 0.6 and 0.2 to $\alpha_1$ and $\alpha_2$ respectively. From Table \ref{Table:contratable}, we see that combining the contradiction probability of NLI doesn't yield consistently better results than the proposed approach across the datasets.

\begin{align*}
\vspace*{-2mm}
\textrm{P}_{prob}  &:= \alpha_1 \textrm{P}_{entail}  + (1-\alpha_1) \textrm{P}_{contradiction} \\
\textrm{P}_{weighted} &:= \alpha_2 \textrm{P}_i  + (1-\alpha_2) \textrm{P}_{prob} \tag{5}
\label{eq:weightedcontra}
\end{align*} \\
Where $\textrm{P}_{entail}$ and $\textrm{P}_{contradiction}$ denote the entailment and contradiction probabilities from NLI and $\textrm{P}_{i}$ refers to the model probability.

\subsection{Demonstration Examples}
\begin{table*}[t!]
\small
\centering

\begin{tabular}{| p{2.8cm}|p{13.5cm}|}
\hline
  \textbf{Article}   & Diego Simeone's side surrendered a 15-game unbeaten run as the Brazilian fired home an 88th-minute volley. A draw would have seen Atletico climb above Barcelona to the summit going into the 10-day winter break. But they had to hold on after losing skipper Gabi Fernandez to two yellows cards in the space of five minutes early in the second half. In a sometimes bad-tempered encounter, the visitors were up against it from the moment Fernandez was cautioned for a foul, then ordered in the 56th minute after committing a handball near the halfway line. Charles saw two first-half chances saved by Atletico keeper Jan Oblak in a game short on clear openings. But the Brazilian had the final say, finding the net with the aid of a deflection off defender Diego Godin for his sixth goal of the season. Barcelona have a game in hand after being without a domestic fixture due to their involvement in the Club World Cup final against River Plate. \\ \hline
     
\textbf{Gold Summary} & Atletico Madrid missed a chance to go top of La Liga after falling to a late winner from Malaga striker Charles.\\ \hline
\textbf{Vanilla beam search} & \colorbox{red} {Cristiano Ronaldo} scored the only goal of the game as Atletico Madrid came from behind to \colorbox{red}{draw 1-1 with 10-man Charles.} \\ \hline
\textbf{Pinocchio} & \colorbox{red}{No summary}\\ \hline
\textbf{FactEdit} & \colorbox{red}{Cristiano Ronaldo} scored the only goal of the game as Atletico Madrid came from behind to \colorbox{red}{draw 1-1 }with Charles in La Liga.
\\ \hline
\textbf{Ours (Saliency V1)} & Atletico Madrid's La Liga play-off hopes suffered a setback as a late goal by \colorbox{red}{Neymar} saw them \colorbox{red}{draw} with Charles.
 \\ \hline
\textbf{Ours (Saliency V2)} & Atletico Madrid's La Liga title hopes suffered a blow as they lost 1-0 to Charles at the Nou Camp. \\ \hline

\end{tabular}

\label{table:halexample}
\end{table*}

\begin{table*}[t!]
\small
\centering
\begin{tabular}{| p{2.8cm}|p{13.5cm}|}
\hline
  \textbf{Article}   & Although there is some common ground between the two governments on, for example, the need for free trade within the single market, Carwyn Jones has complained that he didn't see the letter before it was published on Wednesday. (He has that in common with most of Mrs May's cabinet). The first minister told AMs:  "I discussed the Article 50 letter in general terms with the prime minister when we met in Swansea last week. "I should be clear, though, that I didn't see the letter before today and we were not invited to contribute to its drafting. This is unacceptable and is the culmination of a deeply frustrating process in which the devolved administrations have persistently been treated with a lack of respect. "It is all the more regrettable given the UK government's stated aim was to develop a negotiating framework for the whole of the UK. "Mr Jones may have been playing to an audience, but Welsh Secretary Alun Cairns hit back:  "I'm a bit disappointed in that. The prime minister has been in Wales three times in the last six weeks. "We've been talking about the contents of this letter for many months. "We've clearly all made our representations but, ultimately, the UK government needs to act in the interests of the whole of the UK and that's what we're doing, specifically with Wales being mentioned. " Mrs May did indeed mention Wales in the letter. She told Donald Tusk:  "When it comes to the return of powers back to the United Kingdom, we will consult fully on which powers should reside in Westminster and which should be devolved to Scotland, Wales and Northern Ireland. "But it is the expectation of the government that the outcome of this process will be a significant increase in the decision-making power of each devolved administration. " That sentence may have been written more with Scotland in mind, but it does prompt the question: which powers? Farming? Economic aid? And will the money follow the powers? Alun Cairns wouldn't answer those questions, although Carwyn Jones has said he fears there won't be any money to accompany the powers after 2020. (Perhaps Mr Jones doesn't think Jeremy Corbyn will win power that year - Labour has pledged to maintain EU funding levels on regional aid beyond 2020). Some in Whitehall think the way EU money has been spent in Wales - check out Nick Clegg's film from Ebbw Vale - is an argument for transferring those powers to Westminster, but that looks politically less likely now. We may get some more details about the process in a white paper on the Great Repeal Bill on Thursday but, at the moment, the identity of the powers the UK government wants to see devolved is something of a mystery.\\ \hline
     
\textbf{Gold Summary} & Theresa May's letter triggering Article 50 may have attempted a more conciliatory tone but it does not seem to have worked with the Welsh Government.\\ \hline
\textbf{Vanilla beam search} & \colorbox{red}{Theresa May's letter to the prime minister} is "unacceptable" and "disappointing", according to the first minister.\\ \hline
\textbf{Pinocchio} & \colorbox{red}{No summary}\\ \hline
\textbf{FactEdit} & \colorbox{red}{Theresa May's letter to the prime minister} is "unacceptable" and "disappointing", according to the first minister.
\\ \hline
\textbf{Ours (Saliency V1)} & Theresa May's letter to the Welsh secretary outlining her plans to leave the EU has been criticized by the Welsh Secretary.
 \\ \hline
\textbf{Ours (Saliency V2)} & Prime Minister Theresa May has been criticized by the first minister after the publication of a letter from the prime minister outlining her plans for devolution. \\ \hline

\end{tabular}
\label{table:halexample2}
\end{table*}

\begin{table*}[t!]
\small
\centering

\begin{tabular}{| p{2.8cm}|p{13.5cm}|}
\hline
  \textbf{Article}   & It comes after Camden residents tried to save the St James' Gardens site, close to London Euston, which was a burial ground from 1790 until 1853.Local church warden Dorothea Hackman said it was "quite outrageous" they were going to "dig up our dead".HS2 Ltd said the work would be done with "dignity, respect and care". Notable people buried in the gardens include Capt Matthew Flinders, the first person to circumnavigate and name Australia, and Bill Richmond, one of the first black boxers."[These people] shouldn't be disturbed by spurious activities like this," said Ms Hackman, who helped organise the service, which was expected to be attended by 40 people. "And just think of the detrimental effect removing the benefit of the trees and green space will have on the area in terms of air quality. "There has not been destruction on this scale since the Sixties. Government has run roughshod over democracy. "Resident Marian Kamlish, 92, said that "in times of austerity", such a "vanity project" was an insult to those who work for the likes of the NHS, fire and police forces.HS2 Ltd will excavate sections of the burial ground to enable it to plan the removal of the remains prior to their subsequent re-interment elsewhere. A spokesman stressed the grounds had not been in use for more than a century. "We will ensure that we treat the site with dignity, respect and care," he said. "As such, we will continue to work closely with the local community, the Archbishops' Council, the local parish, Historic England and other organisations as we proceed with the next phase of the project. "In February, Parliament granted powers to build Phase 1 of the line - between London and Birmingham - which is due to open in December 2026.In June, the Department for Transport (DfT) said First Trenitalia West Coast, MTR West Coast Partnership and West Coast Partnership had all been shortlisted to operate the service on the line. \\ \hline
     
\textbf{Gold Summary} & A memorial service has been held for 60,000 people whose remains are due to be exhumed in London as part of the 7bn HS2 high-speed rail project.\\ \hline
\textbf{Vanilla beam search} & Hundreds of people have attended a service to mark the centenary of the \colorbox{red}{birth of the first black boxer.}\\ \hline
\textbf{Pinocchio} & A service has been holds at a Camden church for those buried in a former burial ground to be removed for the HS2 railway line project.\\ \hline
\textbf{FactEdit} & Hundreds of people have attended a service to mark the centenary of the \colorbox{red}{birth of the first black boxer.}
\\ \hline
\textbf{Ours (Saliency V1)} &  A memorial service has been held for people buried in a former cemetery in south-east Britain.
 \\ \hline
\textbf{Ours (Saliency V2)} &  A memorial service has been held in the grounds of a Victorian burial ground in south-east London. \\ \hline

\end{tabular}
\caption{Examples demonstrations, from XSum dataset, showing how the proposed method avoids hallucinations}
\label{table:halexample3}
\end{table*}

\end{document}